\documentclass[conference]{IEEEtran}
\IEEEoverridecommandlockouts
\usepackage{cite,hyperref}
\usepackage{amsmath,amssymb,amsfonts}
\usepackage{algorithmic}
\usepackage{algorithm}
\usepackage{tikz}
\usepackage{graphicx}
\usepackage{textcomp}
\usepackage{xcolor}
\def\BibTeX{{\rm B\kern-.05em{\sc i\kern-.025em b}\kern-.08em
    T\kern-.1667em\lower.7ex\hbox{E}\kern-.125emX}}
\begin{document}

\title{Clinical QA 2.0- Multi-Task Learning for Answer Extraction and Categorization\\
}

\author{\IEEEauthorblockN{1\textsuperscript{st} Priyaranjan Pattnayak}
\IEEEauthorblockA{\textit{University of Washington} \\
Seattle \\
 ppattnay@uw.edu
}\\
\IEEEauthorblockN{2\textsuperscript{nd} Hitesh Patel}
\IEEEauthorblockA{\textit{New York University} \\
New York \\
hitesh.patel945@gmail.com
}
\and

\IEEEauthorblockN{3\textsuperscript{rd} Amit Agarwal}
\IEEEauthorblockA{\textit{Liverpool John Moores University} \\
Liverpool \\
amit.pinaki@gmail.com 
}\\

\IEEEauthorblockN{4\textsuperscript{th} Srikant Panda}
\IEEEauthorblockA{\textit{Birla Institute of Technology} \\
Pilani \\
srikant86.panda@gmail.com
}
\and

\IEEEauthorblockN{5\textsuperscript{th} Bhargava Kumar}
\IEEEauthorblockA{\textit{Columbia University} \\
New York \\
bhargava1409@gmail.com
}\\
\IEEEauthorblockN{6\textsuperscript{th} Tejaswini Kumar}
\IEEEauthorblockA{\textit{Columbia University} \\
New York \\
tejaswinikumar.research@gmail.com
}
}

\maketitle

\begin{abstract}
Clinical Question Answering (CQA) plays a crucial role in medical decision-making, enabling physicians to extract relevant information from Electronic Medical Records (EMRs). While transformer-based models such as BERT, BioBERT, and ClinicalBERT have demonstrated state-of-the-art performance in CQA, existing models lack the ability to simultaneously extract and categorize extracted answers, which is critical for structured retrieval, content filtering, and medical decision support. 

To address this limitation, we introduce a Multi-Task Learning (MTL) framework that jointly trains CQA models for both answer extraction and medical categorization. In addition to predicting answer spans, our model classifies responses into five standardized medical categories: \textbf{Diagnosis, Medication, Symptoms, Procedure, and Lab Reports}. This categorization enables more structured and interpretable outputs, making clinical QA models more useful in real-world healthcare settings.

We evaluate our approach on emrQA, a large-scale medical QA dataset. MTL improves answer extraction F1-score by 2.2\% over ClinicalBERT fine-tuned for QA and achieves a 6.2\% accuracy improvement over ClinicalBERT fine-tuned for classification. These findings suggest that MTL not only enhances CQA performance but also introduces an effective mechanism for categorization and structured medical information retrieval.
\end{abstract}


\section{Introduction}
The rapid digitization of healthcare records has led to an explosion of unstructured textual data in Electronic Medical Records (EMRs) \cite{rasmy2021medbert}. Physicians and healthcare professionals frequently need to extract patient-specific information from these records, which often consist of lengthy and heterogeneous clinical narratives. Clinical Question Answering (CQA) has emerged as a powerful Natural Language Processing (NLP) technique that enables automated retrieval of relevant information from EMRs \cite{lehman2023generalization}. However, deploying QA models in real-world healthcare settings presents significant challenges, including the need for domain adaptation \cite{Pattnayak2017Rainfall}, interpretability, and structured information retrieval.

\subsection{Challenges in Clinical Question Answering}
Despite advancements in deep learning-based QA systems, current approaches struggle with several key challenges:

\begin{itemize}
    \item \textbf{Unstructured Nature of EMRs}: Clinical notes are free-text narratives with inconsistent styles \cite{agarwal2024enhancing}, lacking structured formatting \cite{wagner2022towards}, which hinders accurate QA and necessitates post-processing.

    \item \textbf{Domain Adaptation Limitations}: General models like BERT \cite{devlin2019bert} underperform on medical text. Domain-specific models (e.g., BioBERT \cite{lee2020biobert}, ClinicalBERT \cite{alsentzer2019clinicalbert}, PubMedBERT \cite{gu2021domain}) improve performance but still struggle with medical jargon and abbreviations.

    \item \textbf{Lack of Answer Categorization}: QA models return raw answers without labeling them as \textbf{medication, diagnosis, symptom, procedure}, or \textbf{lab}, limiting EMR integration \cite{zhang2022can} and clinical usability. For example, when asked, \textit{"What medications has the patient been prescribed?"}, a standard QA model might return \textit{"Metformin 500 mg, daily"}.

    \item \textbf{Scalability and Generalization}: Current QA models need extensive fine-tuning, reducing their generalizability across institutions, patient groups, and EMR systems.
\end{itemize}

\subsection{Proposed Approach: Multi-Task Learning for Clinical QA}
To address these challenges, we introduce a Multi-Task Learning (MTL) framework that extends ClinicalBERT beyond a standard classification head by jointly learning answer extraction and medical categorization. This adaptation represents a novel enhancement to ClinicalBERT, enabling it to integrate structured medical knowledge while improving its interpretability and performance. Our approach provides three key contributions:

\begin{enumerate}
    \item \textbf{Multi-Task Learning for Clinical QA}: We enhance existing QA models by adding an auxiliary classification head that categorizes extracted answers into five key medical categories: \textbf{Diagnosis, Medication, Symptoms, Procedures, and Lab Reports}. This allows structured answer retrieval, making it easier for physicians to interpret results.
    
    \item \textbf{Standardized Medical Categorization}: Unlike traditional QA systems that return free-text answers, our method incorporates medical entity recognition using the Unified Medical Language System (UMLS) \cite{bodenreider2004unified}. By aligning extracted answers with standardized medical concepts, our approach ensures interoperability with clinical decision-support tools.

    \item \textbf{Empirical Performance Gains}: We evaluate our model on the emrQA dataset \cite{pampari2018emrqa}, a large-scale corpus for medical QA. Our results demonstrate that the MTL improves answer extraction F1-score by 2.2\% over ClinicalBERT fine-tuned for QA and achieves a 6.2\% accuracy improvement over ClinicalBERT fine-tuned for classification. This suggests that integrating medical categorization with QA enhances both performance and usability in real-world clinical settings.

\end{enumerate}

    
    
    

\subsection{Applications and Impact}
The proposed MTL-based clinical QA model supports several practical use cases:
\begin{itemize}
    \item \textbf{Structured EMR Search}: By categorizing answers, the model enables structured retrieval. For example, a query like \textit{"Show me all past diagnoses"} returns a clean list, bypassing unstructured note scanning.
    
    \item \textbf{Clinical Decision Support}: Categorized outputs can feed into decision-support systems, helping physicians filter relevant medical information and make informed choices.
    
    \item \textbf{Medical Coding and Billing}: Our approach aids in mapping extracted clinical concepts to standardized billing codes (e.g., ICD-10), reducing manual effort for healthcare staff.
    
    \item \textbf{Content Moderation in Clinical AI}: Categorization and rule-based filtering allow AI assistants to flag inappropriate or unsafe responses, such as unlicensed diagnoses or medication suggestions, ensuring regulatory compliance.
\end{itemize}

\section{Related Work}

\subsection{Question Answering in NLP}
Question Answering (QA) is a core NLP task that enables retrieving precise answers from structured and unstructured text \cite{jurafsky2023speech,cahyawijaya2025crowdsource,agarwal2021evaluate}. Early models relied on rule-based techniques and information retrieval, as seen in IBM's Watson \cite{ferrucci2010building}, which matched keywords using structured knowledge bases but lacked deep language understanding. The rise of neural architectures, including Long Short-Term Memory (LSTM) networks and attention mechanisms, significantly improved contextual comprehension.

With Transformer-based models QA took a big leap and \textbf{BERT} \cite{devlin2019bert} set new benchmarks \cite{agarwal2024mvtamperbench}. Fine-tuned variants such as BERT-QA outperformed previous models on datasets like SQuAD \cite{rajpurkar2018know}. Further optimizations, including \textbf{ALBERT} \cite{lan2020albert}, \textbf{RoBERTa} \cite{liu2019roberta}, and \textbf{T5} \cite{raffel2020exploring}, enhanced transformer models for QA across various domains \cite{pattnayak2024survey}. However, their reliance on general-domain corpora limits effectiveness in specialized fields like medicine.

\subsection{Domain Adaptation in Medical NLP}
Medical NLP poses challenges due to domain-specific terminology, abbreviations, and inconsistent documentation \cite{agarwal-etal-2025-fs,yin2024continuous,olaleye2025pseudo}. Standard BERT-based models, trained on general corpora, struggle with these complexities. To address this, domain-adapted models such as \textbf{BioBERT} \cite{lee2020biobert}, pretrained on PubMed abstracts, improved biomedical Named Entity Recognition (NER) and relation extraction \cite{Pattnayak2017AutoSales}. \textbf{ClinicalBERT} \cite{alsentzer2019clinicalbert}, fine-tuned on clinical notes, enhanced hospital-based applications.

\textbf{PubMedBERT} \cite{gu2021domain} eliminated domain mismatches by pretraining exclusively on PubMed abstracts. \textbf{GatorTron} \cite{yang2022large} further improved performance by leveraging de-identified clinical records, while \textbf{Med-BERT} \cite{huang2021clinical} incorporated structured EHR data, bridging the gap between structured and unstructured text.

Despite these advancements, most domain-specific models are fine-tuned separately for tasks like QA, named entity recognition, and classification \cite{agarwal2024domain}. This single-task approach limits generalization and increases data requirements, reducing adaptability in clinical settings.

\subsection{Multi-Task Learning in NLP and Medical AI}
Multi-Task Learning (MTL) has emerged as a powerful technique for improving NLP models by jointly learning related tasks \cite{ruder2017overview}. It has been widely explored in domains such as Named Entity Recognition (NER), Part-of-Speech (POS) tagging, sentiment analysis with syntactic parsing \cite{sogaard2016deep}, and question answering integrated with textual entailment \cite{clark2019boolq}. By leveraging shared representations, MTL enhances generalization and efficiency, especially in data-scarce environments.

In medical AI, MTL has been applied to clinical event detection \cite{li2020multi}, adverse event detection \cite{wei2022multi}, and patient outcome prediction \cite{yoon2022clinical}, consistently outperforming single-task models \cite{liu2022mtlhealth}. For instance, multi-task transformers for clinical risk prediction \cite{mohamed2023riskprediction} have demonstrated improved generalization across patient cohorts. Despite these successes, MTL remains underutilized in Clinical QA, where models predominantly focus on answer extraction while neglecting medical entity classification—critical for structured EMR retrieval and decision support.

\subsection{Limitations and Contribution}
Despite advancements in QA and domain-adapted transformers, key limitations persist: (1) QA models generate raw text outputs, making structured integration into clinical workflows difficult; (2) existing models either extract answers or classify entities, lacking a unified approach; (3) fine-tuned models struggle with distribution shifts in diverse EMR
datasets. To address these gaps, we introduce an MTL framework for Clinical QA that (1) jointly learns answer extraction and medical classification, improving interpretability, (2) integrates domain-adapted transformers like ClinicalBERT with an auxiliary classification head, and (3) enhances generalization by leveraging shared representations, making it more robust for real-world EMR applications.

\section{Dataset and Preprocessing}
\subsection{emrQA Dataset}
We use the emrQA dataset \cite{pampari2018emrqa}, a large-scale corpus designed for Clinical Question Answering (QA) over Electronic Medical Records (EMRs). emrQA was derived from the i2b2 challenge datasets and provides approximately 455,837 QA pairs generated from structured and unstructured clinical documents. 

Unlike traditional QA datasets, emrQA includes both direct retrieval questions (e.g., "What is the patient’s blood pressure?") and inferential questions (e.g., "What medications should be monitored for this patient?").  However, the original dataset does not explicitly label each QA pair with a standardized medical category. To bridge this gap, we apply Named Entity Recognition (NER) using UMLS and SciSpacy to categorize extracted answers into one of the five predefined medical categories.

\subsection{Preprocessing for Multi-Task Learning}
Since emrQA was originally structured for span-based QA, we preprocess it for our joint QA and classification framework:
\begin{algorithm}[H]
\caption{Preprocessing Pipeline for Multi-Task Learning}
\label{alg:preprocessing}
\begin{algorithmic}[1]
\REQUIRE emrQA dataset $\mathcal{D} = \{(Q, C, A)\}$ where $Q$ = Question, $C$ = Context, $A$ = Answer span.
\ENSURE Preprocessed dataset with medical categories.

\FOR{each $(Q, C, A) \in \mathcal{D}$}
    \STATE \textbf{Tokenization:} Apply WordPiece tokenization to $(Q, C, A)$.
    \STATE \textbf{Answer Span Extraction:} Identify $A$’s position in $C$.
    \STATE \textbf{Named Entity Recognition (NER):} Extract medical entities from $A$ using SciSpacy + UMLS.
    \STATE \textbf{Assign Medical Category:}
        \STATE \quad \textbf{Diagnosis} if entity $\in$ UMLS Diagnosis Terminology.
        \STATE \quad \textbf{Medication} if entity $\in$ RxNorm.
        \STATE \quad \textbf{Symptoms, Procedure, or Lab Reports} if entity $\in$ respective medical ontology.
    \STATE \textbf{Dataset Splitting:} Partition into $\mathcal{D}_{train}$ (80\%), $\mathcal{D}_{val}$ (10\%), $\mathcal{D}_{test}$ (10\%).
\ENDFOR
\RETURN Preprocessed dataset $\mathcal{D}_{final} = \{(Q, C, A, \text{Label})\}$
\end{algorithmic}
\end{algorithm}

\begin{itemize}
    \item Tokenization: We apply WordPiece tokenization from BERT to ensure compatibility with transformer-based models.
    \item Answer Span Extraction: Each answer is mapped to its exact position within the clinical text for supervised learning.
    \item Named Entity Recognition (NER) Labeling: We use the Unified Medical Language System (UMLS) \cite{bodenreider2004unified, patel2024llm} and SciSpacy for entity recognition. Extracted answers are assigned one of five medical categories: diagnosis, medication, symptoms, procedures, or lab reports.
    \item Handling Class Imbalance: As shown in Table \ref{tab:dataset_statistics}, some categories (e.g., diagnoses) contain significantly more QA pairs than others (e.g., lab reports). We apply class-weighted loss during training and perform oversampling in underrepresented categories to mitigate imbalance.
    \item Dataset Splitting: The dataset is partitioned into 80\% training, 10\% validation, and 10\% test, using stratified sampling, for unbiased evaluation.
\end{itemize}

\subsection{Dataset Statistics and Class Distribution}
Table \ref{tab:dataset_statistics} presents the dataset distribution across different medical categories along with the distribution of entities extracted via SciSpacy and UMLS.

\begin{table}[h]
\centering
\caption{emrQA Dataset Statistics with Category and UMLS Semantic Types}
\label{tab:dataset_statistics}
\resizebox{\columnwidth}{!}{%
\begin{tabular}{|c|c|c|c|}
\hline
\textbf{Category} & \textbf{QA Pairs} & \textbf{Unique Entities} & \textbf{UMLS Semantic Types (TUI)} \\
\hline
Diagnoses & 141,243 & 8,912 & T047, T019, T033 \\
Medications & 255,908 & 6,548 & T200, T109, T121 \\
Procedures & 20,540 & 1,247 & T060, T061, T058 \\
Symptoms & 23,474 & 1,638 & T184 \\
Lab Reports & 14,672 & 987 & T034, T059 \\
\hline
\textbf{Total} & 455,837 & 19,332 & - \\
\hline
\end{tabular}%
}
\end{table}


The entity distribution reflects the number of unique medical entities identified within each category, highlighting the diversity and granularity of the extracted clinical terms.

\subsection{Example Data Format and Tokenized Representation}
Table \ref{tab:example_data} illustrates an example input format, showing how a clinical QA pair is structured along with tokenized representation.

\begin{table}[h]
\centering
\caption{Example Data Format for QA and Classification}
\label{tab:example_data}
\begin{tabular}{|c|p{5cm}|}
\hline
Field & Example \\
\hline
Question & What medication was prescribed for hypertension? \\
\hline
Context (EMR Text) & The patient was diagnosed with hypertension and prescribed Lisinopril 10mg daily. \\
\hline
Answer (Span) & Lisinopril 10mg daily \\
\hline
Classification Label & Medication \\
\hline
Tokenized Input & \texttt{[CLS] What medication was prescribed for hypertension? [SEP] The patient was diagnosed with hypertension and prescribed Lisinopril 10mg daily. [SEP]} \\
\hline
\end{tabular}
\end{table}

\subsection{Justification for Using emrQA}
The emrQA dataset is particularly well-suited for this study for the following reasons:

\begin{itemize}
    \item Multi-task potential: The dataset structure supports both answer span extraction and medical classification, making it ideal for MTL.
    \item Clinical authenticity: emrQA is derived from real clinical notes rather than synthetic data, ensuring relevance to real-world applications.
    \item Diverse question types: The dataset contains both retrieval-based and inference-based questions, allowing models to handle factual lookups and complex reasoning.
\end{itemize}

By structuring emrQA for joint answer extraction and medical classification, we enable a more interpretable and structured clinical QA framework.

\section{Model Architecture}

\subsection{Overview}
The proposed model extends transformer-based Clinical Question Answering (QA) models by integrating Multi-Task Learning (MTL) to enhance structured information retrieval. The model simultaneously performs:

\begin{itemize}
    \item \textbf{Answer Extraction}: Identifying the most relevant answer span within the clinical text.
    \item \textbf{Medical Entity Classification}: Categorizing the extracted answer into predefined medical categories: Diagnosis, Medication, Symptoms, Procedures, and Lab Reports.
\end{itemize}

We adopt a ClinicalBERT-based architecture with 12 transformer layers, 768 hidden units per layer, and 110M trainable parameters. The model consists of:

\begin{itemize}
    \item A shared transformer encoder that produces contextualized token representations.
    \item A span prediction head (2 fully connected layers, \textit{512, 256 units}, ReLU activation) for answer extraction.
    \item A classification head (3 fully connected layers, \textit{512, 256, 5 units}, Softmax activation) for medical entity prediction.
\end{itemize}

Both heads share lower-layer representations while upper layers are task-specific, reducing parameter redundancy and improving multi-task efficiency. By jointly training these tasks, the model improves both answer relevance and interpretability in clinical decision-support systems.

\subsection{Transformer Model Selection}
We evaluate two domain-specific transformer models:

\begin{itemize}
    \item BioBERT \cite{lee2020biobert}, pretrained on biomedical literature, optimized for general biomedical NLP tasks.
    \item ClinicalBERT \cite{alsentzer2019clinicalbert}, trained on MIMIC-III clinical notes, suitable for processing electronic health records (EHRs).
\end{itemize}

ClinicalBERT is the primary model due to its alignment with clinical narratives. It captures domain-specific terminology, abbreviations, and structured note-writing styles found in EHRs.  

BioBERT, which is trained on biomedical literature from PubMed and PMC, serves as a secondary baseline to assess generalization performance. BioBERT is not exposed to real-world clinical text, which often contains informal phrasing and abbreviations unique to doctor-patient interactions. This comparison allows us to evaluate the impact of domain-specific pretraining when applied to QA tasks involving real-world clinical notes.

\subsection{Multi-Task Learning Setup}
The transformer encoder produces contextualized embeddings that are passed to two task-specific heads: 1) A span prediction module for answer extraction, 2) A classification module for medical entity categorization.


\begin{figure}[h]
    \centering
    \includegraphics[width=0.48\textwidth]{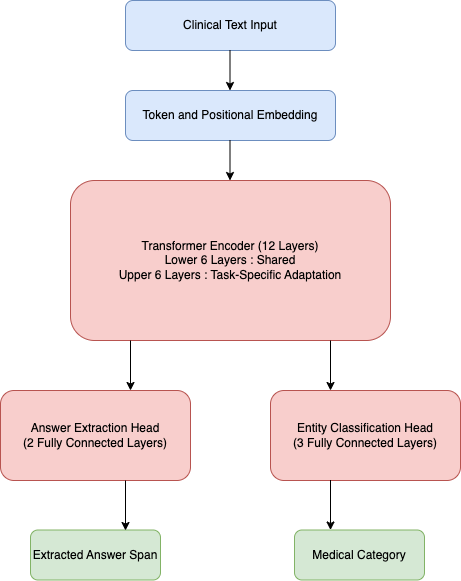}
    \caption{Architecture of the Multi-Task Clinical QA model. A 12-layer transformer encoder shares lower layers across tasks, while upper layers specialize in Answer Extraction and Medical Entity Classification.}
    \label{fig:multiclass_architecture}
\end{figure}

\subsection{Handling Task Interference}
To prevent gradients from one task from negatively affecting the other, we implement layer-wise attention masking, a technique inspired by task-specific adapter layers \cite{Stickland2019BAM, Houlsby2019Adapter}. 

In this setup:
\begin{itemize}
    \item The lower 6 layers of the transformer remain shared, capturing general clinical text representations.
    \item The upper 6 layers are task-specific, with independent attention mechanisms for QA and classification.
    \item Attention masking is applied at the task-specific layers to prevent cross-task gradient interference, ensuring that QA-specific gradients do not distort classification representations.
\end{itemize}

This mechanism ensures that shared lower layers generalize across both tasks, while upper layers specialize in their respective tasks, mitigating negative transfer effects in multi-task learning \cite{Ruder2017MTL}.



\subsection{Handling Answer Ambiguity}
In clinical settings, extracted answers may belong to multiple medical categories. For instance, "Metformin 500mg" can be classified as both Medication (a prescribed drug) and Lab Report (if extracted from test results). 

To address this, we adopt soft classification labels, a technique widely used in multi-label classification \cite{Bengio2015SoftLabels, Joulin2017Softmax}. Instead of a single category per answer, the model assigns probabilities to multiple categories. This is achieved using sigmoid activation rather than softmax:

\begin{equation}
    P(y|x) = \sigma(W_h f(x) + b)
\end{equation}

where \( f(x) \) is the feature representation from ClinicalBERT, and \( \sigma \) ensures multi-label assignments. During training, we minimize Binary Cross-Entropy Loss (BCE Loss), allowing the model to handle ambiguous classifications effectively.

\subsection{Loss Function and Optimization}

A weighted loss function is used to balance both the QA and classification tasks while addressing class imbalance:

\begin{equation}
    \mathcal{L} = \lambda_{\text{QA}} \mathcal{L}_{\text{QA}} + \lambda_{\text{Class}} \sum_{i} w_i \mathcal{L}_{\text{CE}}(y_i, \hat{y}_i),
\end{equation}

where:
\begin{itemize}
    \item \( \mathcal{L}_{\text{QA}} \) is the negative log-likelihood loss for answer span prediction.
    \item \( \mathcal{L}_{\text{Class}} \) is the category-weighted cross-entropy loss for classification.
    \item \( w_i \) is the class-specific weight, computed as:
    \begin{equation}
        w_i = \frac{N}{|C| \cdot \text{count}(i)}
    \end{equation}
    where \( N \) is the total number of QA pairs, \( |C| \) is the number of categories, and \( \text{count}(i) \) represents the number of examples in category \( i \).
    \item \( \lambda_{\text{QA}} \) and \( \lambda_{\text{Class}} \) control the contribution of each task.
\end{itemize}

The values of \( \lambda_{\text{QA}} \) and \( \lambda_{\text{Class}} \) are selected using grid search, optimizing for both QA F1-score and classification accuracy. For class imbalance, we compute category-specific weights \( w_i \) and apply oversampling in underrepresented categories.

\subsection{Training and Evaluation Metrics}
The model is trained using AdamW with a learning rate of $5 \times 10^{-5}$ for 5 epochs. The training process includes:

\begin{itemize}
    \item Learning rate warm-up for the first 10\% of training steps, followed by linear decay.
    \item Early stopping based on validation loss to prevent overfitting.
\end{itemize}

We select evaluation metrics tailored to the dual-task nature of the model:

\begin{itemize}
    \item F1-score for answer span extraction, as it balances precision and recall, which is critical in medical QA where partial answers may still be relevant.
    \item Exact Match (EM) to measure strict correctness in extracted spans. While this is a standard metric for QA, it is less suitable for medical QA since minor phrasing differences (e.g., "Tylenol 500mg" vs. "Tylenol") may still be clinically correct.
    \item Accuracy and weighted F1-score for classification. Since medical entity categories are imbalanced (e.g., more diagnoses than lab results), weighted F1 prevents minority classes from being ignored.
\end{itemize}

F1-score is prioritized over accuracy for answer extraction because, in medical settings, the model should aim to maximize recall (retrieving clinically important information) while maintaining precision.

\subsection{Algorithm: Multi-Task Learning for Clinical QA}
The training process is formalized in Algorithm \ref{alg:mtl_qa}.

\begin{algorithm}[h]
\caption{Multi-Task Learning for Clinical QA}
\label{alg:mtl_qa}
\begin{algorithmic}[1]
\REQUIRE Transformer model (ClinicalBERT), training dataset $\mathcal{D}$
\STATE Initialize model parameters $\theta$
\FOR{each epoch}
    \FOR{each batch $(q, d, a, y) \in \mathcal{D}$}
        \STATE Compute transformer embeddings: $\mathbf{H} \leftarrow \text{Transformer}(q, d)$
        \STATE Predict answer span $(P_{\text{start}}, P_{\text{end}})$
        \STATE Predict classification label: $P_{\text{class}}$
        \STATE Compute multi-task loss:
        \STATE $\mathcal{L} = \lambda_{\text{QA}} \mathcal{L}_{\text{QA}} + \lambda_{\text{Class}} \mathcal{L}_{\text{Class}}$
        \STATE Update parameters using AdamW
    \ENDFOR
\ENDFOR
\RETURN Optimized model parameters $\theta$
\end{algorithmic}
\end{algorithm}

\section{Experimental Results}
\subsection{Overview}
We evaluate the performance of our Multi-Task Learning (MTL) framework on the emrQA dataset using two domain-specific transformer models: BioBERT and ClinicalBERT. We measure performance on two tasks: answer span extraction and medical entity classification.

\subsection{Performance Comparison: BioBERT vs. ClinicalBERT}
Table \ref{tab:model_comparison} presents the performance of BioBERT and ClinicalBERT on the two tasks. ClinicalBERT consistently outperforms BioBERT across all metrics due to its pretraining on electronic health records.


\begin{table}[h]
\centering
\caption{Performance Comparison: BioBERT vs. ClinicalBERT}
\label{tab:model_comparison}
\begin{tabular}{lcc|cc}
\hline
Metric & \multicolumn{2}{c|}{Answer Extraction} & \multicolumn{2}{c}{Entity Classification} \\ 
\hline
 & BioBERT & ClinicalBERT & BioBERT & ClinicalBERT \\
\hline
F1-score & 76.3 & \textbf{81.7} & 79.5 & \textbf{85.2} \\
Recall & 74.9 & \textbf{79.8} & 77.2 & \textbf{83.6} \\
Exact Match & 64.2 & \textbf{69.5} & - & - \\
Accuracy & - & - & 82.4 & \textbf{90.7} \\
Weighted F1 & - & - & 78.9 & \textbf{84.7} \\
\hline
\end{tabular}
\end{table}


\subsubsection{Discussion}
The results confirm that ClinicalBERT outperforms BioBERT across all evaluation metrics, particularly in answer extraction (F1-score: 81.7 vs. 76.3) and classification accuracy (90.7 vs. 82.4) as shown in Table \ref{tab:model_comparison}. These improvements stem from ClinicalBERT’s exposure to structured clinical narratives, which helps in recognizing domain-specific terminology and abbreviations.

Higher recall in answer extraction (79.8\% vs. 74.9\%) suggests that ClinicalBERT better identifies relevant spans, but the exact match score remains lower due to paraphrased variations in answers. In entity classification, the weighted F1-score improves by nearly 6\% with ClinicalBERT, reinforcing its advantage in domain-specific medical text processing.

\subsection{Ablation Study: Impact of Multi-Task Learning vs. Standard Fine-Tuning}
To assess the contribution of multi-task learning, we compare it to a standard fine-tuning approach where ClinicalBERT is trained solely on QA or classification without multi-task objectives.

\begin{table}[ht]
\centering
\caption{Ablation Study: MTL vs. Standard Fine-Tuning}
\label{tab:mtl_vs_finetune}
\begin{tabular}{p{3cm}l p{2cm}c p{1.8cm}c}
\hline
Configuration & Answer Extraction (F1) & Classification Accuracy \\
\hline
Fine-Tuned ClinicalBERT (QA only) & 79.5 & NA \\
Fine-Tuned ClinicalBERT (Classification only) & NA & 84.5 \\
Multi-Task Learning (QA + Classification) & \textbf{81.7} & \textbf{90.7} \\
\hline
\end{tabular}
\end{table}

\subsubsection{Discussion}
The multi-task model outperforms standard fine-tuned ClinicalBERT by \textbf{2.2\%} in F1-score for answer extraction and achieves \textbf{90.7\%} accuracy in entity classification as compared to 84.5\% in ClinicalBERT fine-tuned for classification, confirming the effectiveness of MTL. These improvements indicate that shared representations between tasks provide additional contextual cues, leading to better span prediction and more robust entity recognition.

The improved classification accuracy also demonstrates MTL’s potential for content moderation and structured information retrieval, as it refines answer categorization and entity structuring in clinical datasets.

\subsection{Class-Wise Entity Classification Performance}
\begin{figure}[h]
    \centering
    \includegraphics[width=0.48\textwidth]{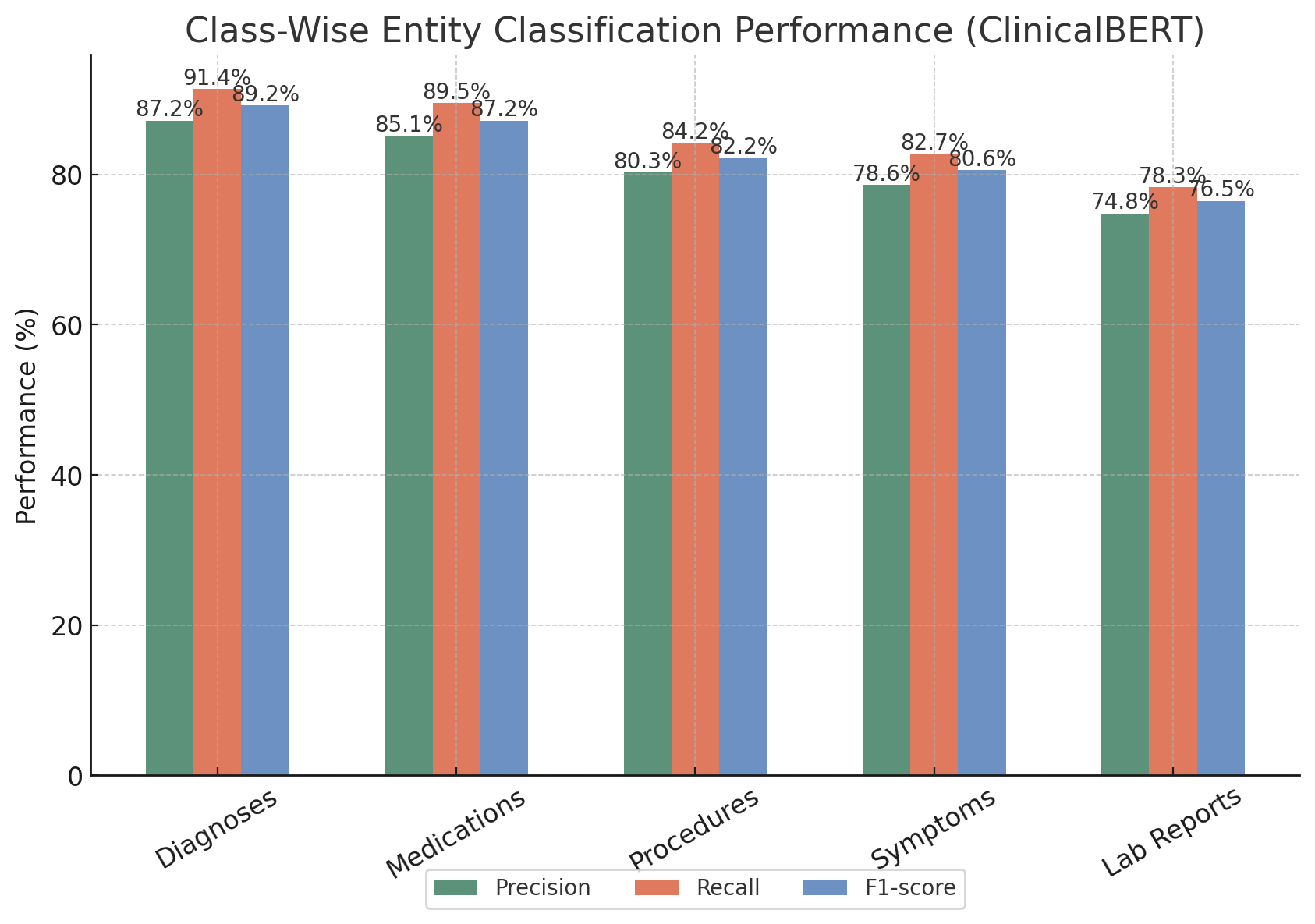}
    \caption{Class-Wise Entity Classification Performance for ClinicalBERT.}
    \label{fig:classwise_performance}
\end{figure}

To further analyze classification performance across medical categories, Fig. \ref{fig:classwise_performance} presents precision and recall per entity type.


\subsubsection{Discussion}
Lab-related entities exhibit lower recall (78.3\%), likely due to inconsistent naming conventions and variations in reporting formats. In contrast, diagnoses and medications achieve the highest precision and recall, suggesting that these categories benefit from clearer contextual patterns in clinical text.

\subsection{Error Analysis}
Table \ref{tab:error_analysis} categorizes common errors in model predictions.

\begin{table}[h]
\centering
\caption{Error Analysis}
\label{tab:error_analysis}
\begin{tabular}{lc}
\hline
Error Type & Frequency (\%) \\
\hline
Ambiguous Answers & 28.3 \\
Span Boundary Errors & 23.1 \\
Incorrect Entity Classification & 18.7 \\
OOV Medical Terms & 15.6 \\
Others & 14.3 \\
\hline
\end{tabular}
\end{table}

\subsubsection{Discussion}
Ambiguous answers (28.3\%) are the most frequent error, often due to multiple valid responses. Span boundary mismatches (23.1\%) highlight the importance of annotation consistency. Incorrect entity classification (18.7\%) suggests that incorporating external medical ontologies, such as UMLS, could further enhance classification accuracy.

\section{Conclusion}
This study introduced a Multi-Task Learning (MTL) framework for Clinical Question Answering (CQA), integrating answer extraction with medical entity classification. Our approach enhances structured information retrieval, improving both accuracy and interpretability in clinical decision support.

Empirical results on emrQA show that MTL boosts QA F1-score by 2.2\% and classification accuracy by 6.2\% over fine-tuned ClinicalBERT for both the tasks respectively. By leveraging shared representations, our model mitigates negative transfer effects, ensuring robust performance across diverse medical queries.

Despite these improvements, challenges remain, including handling paraphrased clinical queries and integrating external medical ontologies such as UMLS for enhanced classification. Future research should explore retrieval-augmented methods and contrastive learning to improve generalization. Expanding to multilingual clinical QA will further enhance applicability across diverse healthcare settings.

Our findings contribute to advancing clinical NLP by bridging unstructured EMR text with structured retrieval, supporting more interpretable and efficient AI-driven medical question answering.

\bibliographystyle{IEEEtran}
\bibliography{custom}



\end{document}